\pdfoutput=1
\documentclass[conference]{IEEEtran}
\IEEEoverridecommandlockouts
\usepackage{url}
\usepackage{hyperref}
\hypersetup{
  colorlinks = true,
  linkcolor  = blue,
  urlcolor   = blue,
  citecolor  = black,
}
\usepackage{placeins}
\usepackage{amsmath}
\usepackage{cleveref}
\newcommand*{\fullref}[1]{\hyperref[{#1}]{\cref*{#1} \nameref*{#1}}}
\newcommand*{\Fullref}[1]{\hyperref[{#1}]{\Cref*{#1} \nameref*{#1}}}
\newcommand*{\secref}[1]{\hyperref[{#1}]{\autoref*{#1}}}
\newcommand*{\Secref}[1]{\hyperref[{#1}]{\Cref*{#1}}}

\let\oldFootnote\footnote
\newcommand\nextToken\relax

\renewcommand\footnote[1]{%
    \oldFootnote{#1}\futurelet\nextToken\isFootnote}

\newcommand\isFootnote{%
    \ifx\footnote\nextToken\textsuperscript{,}\fi}

\usepackage{thmbox}
\usepackage{mdframed}
\usepackage{cite}
\usepackage{amsmath,amssymb,amsfonts}
\usepackage{algorithmic}
\usepackage{graphicx}
\usepackage{textcomp}
\usepackage{tabularx}
\usepackage{booktabs}
\usepackage{xcolor}
\def\BibTeX{{\rm B\kern-.05em{\sc i\kern-.025em b}\kern-.08em
    T\kern-.1667em\lower.7ex\hbox{E}\kern-.125emX}}

\newcounter{openboxwithtitle}[section]
\newenvironment{openboxwithtitle}[1]{
  \refstepcounter{openboxwithtitle} 
  \vspace{1.75ex}
  \thmbox[L]{\textbf{#1}}
  \hspace*{-1.5em}\slshape\ignorespaces%
}{
  \endthmbox\vspace*{.75ex}
}

\begin{document}

\title{Computational Fact-Checking of Online Discourse: Scoring Scientific Accuracy in Climate Change Related News Articles
\thanks{
SE²A Excellence Cluster [EXC 2163], NFDI4Ing project (DFG project number 442146713), NFDI4DataScience (DFG project number 460234259)
}
}

\author{
    \IEEEauthorblockN{
    Tim Wittenborg\IEEEauthorrefmark{1},
    Constantin Sebastian Tremel\IEEEauthorrefmark{1},
    Oliver Karras\IEEEauthorrefmark{2},
    Sören Auer\IEEEauthorrefmark{1}\IEEEauthorrefmark{2}}
    \IEEEauthorblockA{\IEEEauthorrefmark{1}L3S Research Center, Leibniz University Hanover, Hanover, Germany
    \\tim.wittenborg@l3s.uni-hannover.de, constantin.tremel@stud.uni-hannover.de}
    \IEEEauthorblockA{\IEEEauthorrefmark{2}TIB - Leibniz Information Centre for Science and Technology, Hanover, Germany
    \\oliver.karras@tib.eu, soeren.auer@tib.eu}
}

\maketitle

\begin{abstract}
Democratic societies need reliable information.
Misinformation in popular media, such as news articles or videos, threatens to impair civic discourse.
Citizens are, unfortunately, not equipped to verify the flood of content consumed daily at increasing rates.
This work aims to quantify the scientific accuracy of online media semi-automatically.
We investigate the state of the art of climate-related ground truth knowledge representation.
By semantifying media content of unknown veracity, their statements can be compared against these ground truth knowledge graphs.
We implemented a workflow using LLM-based statement extraction and knowledge graph analysis.
Our implementation can streamline content processing towards state-of-the-art knowledge representation and veracity quantification.
Developed and evaluated with the help of 27 experts and detailed interviews with 10, the tool evidently provides a beneficial veracity indication.
These findings are supported by 43 anonymous participants from a parallel user survey. 
This initial step, however, is unable to annotate public media at the required granularity and scale.
Additionally, the identified state of climate change knowledge graphs is vastly insufficient to support this neurosymbolic fact-checking approach.
Further work towards a FAIR (Findable, Accessible, Interoperable, Reusable) ground truth and complementary metrics is required to support civic discourse scientifically.
\end{abstract}

\begin{IEEEkeywords}
Scientific Accuracy, Climate Change, Fake News, LLM
\end{IEEEkeywords}

\section{Introduction}
Over 100 zettabytes of online web content are created, captured, copied, and consumed globally every year~\cite{statista_daten_2023}.
Every minute, humanity watches 43 years of streaming content and sends over 280 million instant messages~\cite{domo}.
The veracity of this content is often unclear to recipients.
Yet, accurate information is vital for political topics like climate action, where a rapidly closing window of opportunity~\cite{lee_ipcc_2023} meets insufficient policies~\cite{emissiongapReport2023} and debates not aligning with the overwhelming scientific consensus~\cite{lee_ipcc_2023,cook2013}.
While traditional fact-checking organizations perform vital work, they are increasingly overwhelmed by the volume and speed of online content dissemination. 
To address this scalability challenge, computational fact-checking offers a promising augmentation, enabling systematic, semi-automated assessments of veracity. 

We present our work towards scoring the veracity of online content through a computational fact-checking pipeline.
Our approach employs a neurosymbolic method for fact validation, combining the linguistic competence of Large Language Models (LLMs) with the structured rigor of knowledge graphs. 
First, an LLM extracts triple statements from unstructured media.
These extracted triples are then normalized and disambiguated.
Subsequently, the statements are matched against a curated ground truth knowledge graph, composed of verified scientific claims, particularly from authoritative sources in the climate change domain. 
This enables a symbolic verification process where the proximity, overlap, or contradiction between the extracted statements and the knowledge graph is quantified.
The resulting scientific accuracy score provides a scalable, interpretable, and traceable indication of content veracity.

Our contributions include:
\begin{enumerate}
    \item An overview of existing and required ground truth knowledge graphs for the climate change domain, alongside current fact-checking support, identifying the potential and limits of computational fact-checking.
    \item A streamlined workflow from ground-truth semantification to equally processing popular media and scoring scientific accuracy, available as a modular open-source implementation\footnote{\url{https://github.com/borgnetzwerk/fact-checking}}.
    \item Two evaluations of this tool from both experts and users, substantiating the necessity and usability, as well as an indication of required future work towards sustainably scaling fact-checking amid the information flood.
\end{enumerate}

The remainder of the paper is structured as follows:\\
\Secref{sec:related} presents related work regarding 
several \nameref{sec:cfc} attempts to automate the process, and an analysis of the therefore required \nameref{sec:kg}.
\Secref{sec:approach} and \secref{sec:implementation} detail our approach towards developing the workflow, while the former focuses on the methodology and the latter on the technical details.
\Secref{sec:evaluation} describes the \nameref{sec:design} and \nameref{sec:results} of our evaluation, including \nameref{sec:validity}.
\Secref{sec:discussion} discusses these findings and motivates future work, while
\secref{sec:conclusion} concludes the paper.

\begin{table*}[t!]
\centering
\caption{Comparison of Related Work in Computational Fact-Checking}
\begin{tabular}{|p{2.4cm}|p{2.4cm}|p{4.4cm}|p{1.3cm}|p{0.8cm}|p{4.2cm}|}
\hline
\textbf{Approach} & \textbf{Data Source} & \textbf{Fact-Checking Method} & \textbf{Domain Focus} & \textbf{Open Source} & \textbf{Limitations} \\
\hline
Ciampaglia et al. (2015) \cite{ciampaglia_computational_2015} & DBpedia & Semantic proximity in knowledge graphs (path length, degree of nodes) & General & Partially & Limited contextual understanding; scalability concerns \\
\hline
Thorne et al. (2018) FEVER \cite{thorne_fever_2018} & Wikipedia & Textual entailment using NLP models; human annotations & General & Yes & Requires structured textual evidence; limited to Wikipedia domain \\
\hline
Leippold et al. (2025) \cite{leippold_automated_2025} & Scientific climate~texts & LLM-based claim verification (black-box) & Climate Change & No & persistent knowledge preservation and process transparency issues \\
\hline
Dessì et al. (2022) SCICERO \cite{dessi_scicero_2022} & Computer~science literature & Deep learning + NER for research knowledge graph generation & Scientific Knowledge & Yes & Domain-specific; not focused on media claim validation \\
\hline
\textbf{This Work (Tremel, 2024)} \cite{tremel_scientific_2024} & Climate reports (e.g., IPCC), online media & LLM-based triple extraction + symbolic verification in KG & Climate Change & Yes & Limited ground truth; scalability and context handling remain challenges \\
\hline
\end{tabular}
\label{tab:related_work_comparison}
\end{table*}

\section{Related work\label{sec:related}}
\subsection{Computational Fact Checking\label{sec:cfc}}
Fact checking experts are often organized as organizations, such as Science Feedback or FactCheck.org, which coordinate guidelines and insights via overlapping meta networks such as 
the International Fact-Checking Network (IFCN), European Digital Media Observatory (EDMO) or European Fact-Checking Standards Network (EFCSN).
They are committed to specific standards for selecting, investigating, reviewing, and publicizing fact-checks, yet vary significantly in the types of content they assess, their rating systems, and the degree of automation.
Ciampaglia et al.~\cite{ciampaglia_computational_2015} state that ``fact checking by expert journalists cannot keep up with the enormous volume of information that is now generated online''.
To mitigate this, Ciampaglia et al. propose computational fact-checking based on semantic proximity, aggregating the generality (degree) of two concept nodes in a knowledge graph by tracing the path between.
Thorne et al.~\cite{thorne_fever_2018} introduce FEVER (Fact Extraction and VERification), a dataset for verification against textual sources.
Leippold et al.~\cite{leippold_automated_2025} demonstrate the fact-checking effectiveness of LLMs on climate claims.
While the results are promising, several black-box issues arise in their process. 
Neither the prompts, code, data, nor exact methodology for PDF parsing and information retrieval are made public. 
This is in addition to the reliance on LLMs' inherent transparency issues.
Dessí et al.~\cite{dessi_scicero_2022} present SCICERO, an open source approach that extracts text from research articles to automatically generate a knowledge graph of research entities.
This improves reusability by storing the processed knowledge in knowledge graphs, an inherently transparent knowledge representation.
The reuse of top- and mid-level ontologies, such as the Basic Formal Ontology (BFO) and Common Core Ontology (CCO), facilitates collective efforts and prevents the duplication of work~\cite{davarpanah_climate_2023}.
\Secref{tab:related_work_comparison} provides an overview of this related work.

\subsection{Ground Truth Knowledge Graphs\label{sec:kg}}
Islam et al.~\cite{islam_knowurenvironment_2022} proposed a knowledge graph for climate change, extracting 411,860 statements from 152,000 domain-relevant scientific articles selected from an 8.1 million article dataset.
Of those 411,860 statements, 24,263 were identified as trusted unique statements and stored in a normalized, but not semantically disambiguated CSV\footnote{\url{https://github.com/saiful1105020/KnowUREnvironment}}.
Yadav et al.~\cite{worthington_semanticclimate_2024} mark these climate knowledge graphs as ``community action priorities'' in the context of the \#semanticClimate initiative\footnote{\url{https://semanticclimate.github.io}}.
They share equitable scientific climate data through open notebook science and citizen engagement, and have developed several tools to liberate information regarding climate change from its PDF confinement, like the Open Climate Reader\footnote{\url{https://semanticclimate.github.io/city-open-climate-reader}}.
Their ClimateKG\footnote{\url{https://github.com/semanticClimate/climate-knowledge-graph}} project just started in 2025, focusing on enhancing large scientific corpora with semantic and linked open data using RDF and Wikibase for open science indexing and cataloguing.
The sources used to construct such ground truth knowledge graphs must be carefully selected, i.e., reproducible, traceable, reputable, and peer-reviewed primary literature.
Despite collaborative efforts such as the IPCC, Climate Change Performance Index (CCPI), Corporate Climate Responsibility Monitor (CCRM), or Science Daily Climate Change (SciDCC) providing extensive knowledge synthesis, the outputs often culminate in PDF format, limiting all aspects of FAIRness~\cite{wilkinson_fair_2016}, especially interoperability.

Current infrastructure remains insufficient to manage the growing volume of (mis-)information.
Despite robust systems such as the Open Research Knowledge Graph (ORKG)~\cite{auer2025open}, essential knowledge is still frequently locked in siloed formats, even in key knowledge infrastructure such as the IPCC.

\begin{figure*}[t!]
  \includegraphics[width=\textwidth]{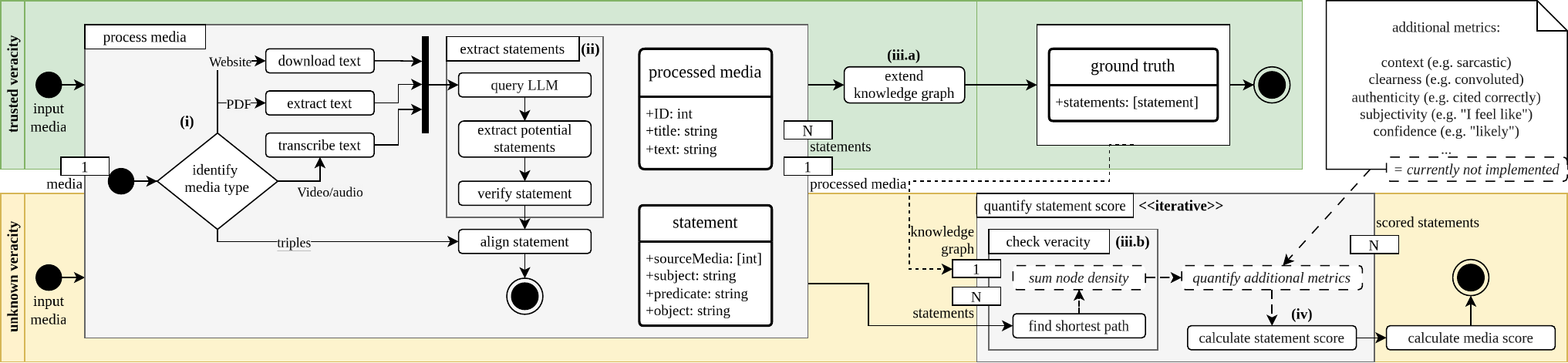}
  \caption{Proposed scoring pipeline consisting of (i) textualizing media from different file types, (ii) LLM-based statement extraction, verification, and alignment. (iii.a) Trusted statements extend the ground truth knowledge graph, (iii.b) untrusted statements are checked for veracity using graph analysis on the ground truth, concluding in (iv) a final score calculation.}
  \label{fig:teaser}
\end{figure*}

\section{Approach\label{sec:approach}}
Our work sets out to answer the following research question:

\begin{openboxwithtitle}{Research Question}
~How can natural language processing and knowledge graphs help quantify the
scientific accuracy of secondary literature in the context of climate change?
\end{openboxwithtitle}

Our approach leverages the interoperability and reusability of climate-knowledge-graph-based fact-checking approaches alongside LLM NLP capabilities.
This aims to mitigate the LLM's transparency and reproducibility limitations with persistent knowledge representations, and enable reusable, scalable computational fact-checking on public media.
\begin{figure}[bt!]
   \centering
    \includegraphics[clip,trim={1cm 1.5cm 1cm 1cm},width=0.45\textwidth]{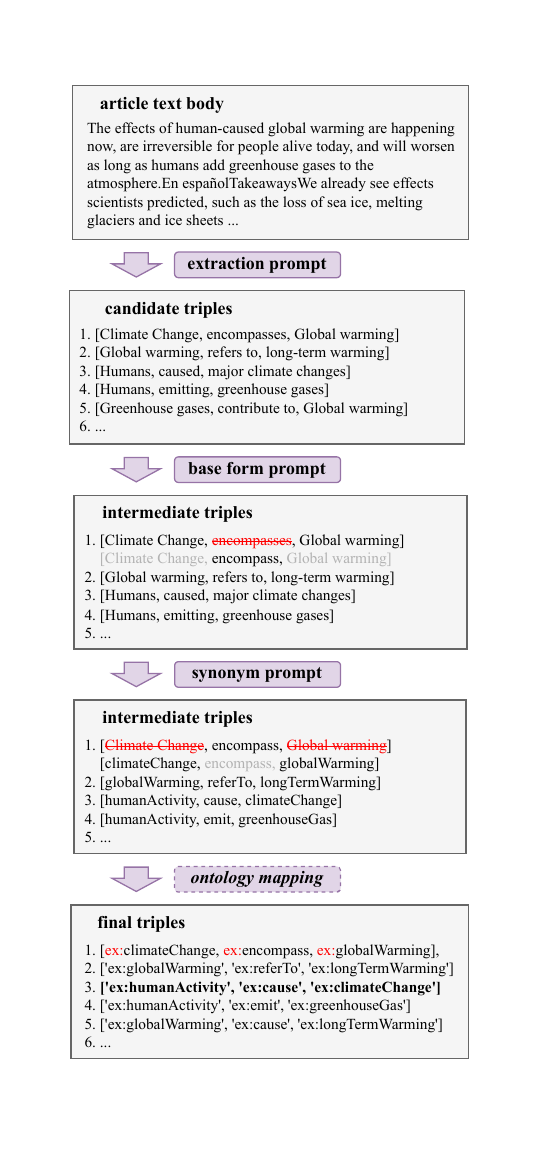}
    \caption[Triple extraction workflow example from text body to aligned triples. An LLM was used to handle initial extraction, base forms and synonyms. A prototypical ontology mapping was implemented to a placeholder example ontology.]{Triple extraction workflow example from text body\footnotemark{} to aligned triples. An LLM was used to handle initial extraction, base forms and synonyms. A prototypical ontology mapping was implemented to a placeholder example ontology.}
    \label{fig:example}
\end{figure}
To develop this approach, we iteratively researched and tested our workflow modules, with intermediate expert meetings for further refinement as well as confirming and potentially overcoming apparent roadblocks.
These experts were most notably from the ORKG team, a group of knowledge graph experts with several members specializing in NLP and neurosymbolic applications.
The latter are further referred to as NLP experts from the ORKG team, while the entire group is called the ORKG team.
The NLP experts were consulted many times during the development, while the entire team was consulted twice, once for an intermediate presentation and once for a closing presentation, with the potential for individual follow-up interviews. 
The pipeline depicted in \secref{fig:teaser} results from these iterations, detailed in the following section.

\section{Implementation\label{sec:implementation}}
This section navigates through the seven modules of the pipeline, from \nameref{sec:process} to \nameref{sec:score}.
\subsubsection{Process Media\label{sec:process}}
This module assigns each retrieved media a unique ID and then processes it into a uniform triple structure, often with the intermediate \nameref{sec:extraction} step for not already structured data.
Various types of media can be processed:
For non-textual documents, such as video and audio files, transcription was facilitated using the OpenAI Whisper base model.
Text from web documents is extracted using the Python library Beautiful Soup, from PDF files using PDFMiner.
Such retrieval methods are noisy (see "En español" in \secref{fig:example}), yet another reason to advance FAIR knowledge.

\subsubsection{Statement Extraction\label{sec:extraction}}
We investigated existing techniques to efficiently extract reliable triple statements at scale, considering context, particularly for the climate domain, but preferably domain-independent. 
Predicate normalization and semantic alignment, such as synonym detection, facilitate a comparable statement structure.
Since our research could not identify an approach that clearly suited our needs, we tested the three most promising approaches. 
First, Semantic Role Labeling (SRL)~\cite{he_deep_2017} was used to create Abstract Meaning Representation (AMR)~\cite{banarescu_abstract_2013} graphs.
Disambiguating unrestricted AMR graphs and using them for semantically consistent and reliable fact-checking was highly time-intensive.
Once set up, these graphs work consistently, but their setup does not scale well to the complexity and number of statements required.
Second, we explored triple extraction using a domain-specific Named Entity Recognition (NER) model.
Training a domain-specific model combined with an ontology could provide better results, but is resource-intensive and unsuitable for the heterogeneous domains of public discourse.
Our third approach utilized Large Language Models (LLMs) to extract and align triples.
While LLMs are currently not reliable either, human-in-the-loop supervision promises faster, sufficiently reliable results.
After iterative testing, these initial assessments were presented to NLP experts from the ORKG team. 
The findings were substantiated, and the LLM-based approach was considered sufficient for producing a proof of concept within the scope.
The LLMs tested for this approach include Mistral-7B-v0.1, Llama-2-7b-chat-hf, and ChatGPT3.5 UI.
Among all models tested using this approach, ChatGPT (GPT3.5) proved to be the most effective at extracting and formatting triples.

\subsubsection{Alignment} 
Predicate normalization and semantic alignment, such as synonym detection, concluded the triple processing.
At this stage of the implementation, this is achieved by prompting the same LLM.
Each triple is assigned a media ID array, allowing the statement origin to be traceable through multiple sources.
Finally, the extracted and aligned triples are saved in Turtle (Terse RDF Triple Language) format.

\subsubsection{Extend Knowledge Graph\label{sec:extendKG}}
If the source was trusted, we ingested the RDF serialized triples into a GraphDB database.
RDF-star and SPARQL-star are available in GraphDB for managing triple metadata and annotations, alonside querying and visualizing the constructed knowledge graph.
For future implementations, GraphDB also offers an API that enables the automation of steps that are currently done via the UI.
Statements about statements enable linking sources, add confidence and objectivity scores for statements, and provide temporal context.  
Since no existing knowledge graph of satisfactory quality existed, we created a prototypical set of triples from the IPCC AR6~\cite{lee_ipcc_2023}, the highly synthesized consensus of climate science.
This PDF data was processed using our approach described in \secref{sec:extraction} and manual validation.
Consideration for an ontology mapping is also prototypically included, providing structure, inferencing potential and interoperability.
An extensive, federated, ontology-backed climate graph ecosystem is advisable for a strong implementation, though currently not available.

\subsubsection{Check Veracity\label{sec:check}}
The veracity check involves knowledge graph analysis, initially searching for exact matches of the proposed triples in the ground truth, using SPARQL queries.
If no exact match is found, a path check according to Ciampaglia et al.~\cite{ciampaglia_computational_2015} could theoretically approximate the veracity by assessing the path length and analyzing the nodes and edges crossed.
Before this check can be used to supplement an estimation and a meaningful judgment in unclear cases, a vastly more extensive ground truth is required, as aforementioned.

\subsubsection{Quantify Additional Metrics\label{sec:additional}}
In addition to veracity, the conceptual score may include other criteria, such as temporal relevance, confidence, clearness, transparency, information depth, objectivity, and rationality.
Neglecting these factors can negatively impact the annotation quality.
Due to the currently limited ground truth knowledge representation data available, no satisfactory computational quantifications could be identified.
Similarly to the ontology mapping, our approach still considers this important, yet currently insufficiently implementable step, and recommends further advancement.

\subsubsection{Calculate Score\label{sec:score}}
The overall accuracy score $s_{\mathrm{acc}} \in [0, 1]$ is calculated using the weighted individual metrics: $s_{\mathrm{acc}} = \sum_{\mathrm{i=1}}^{\mathrm{n}} s_\mathrm{i} \cdot w_\mathrm{i}$, where $s_\mathrm{i} \in [0, 1]$ and $\sum_{\mathrm{i=1}}^{\mathrm{n}} w_\mathrm{i} = 1$. 
According to our investigation of computational fact-checking, accuracy consists mostly of veracity: $w_{\mathrm{ver}} \geqslant 0.5 \geqslant \sum_{\mathrm{i=1,} \ \mathrm{i} \neq \mathrm{ver}}^{\mathrm{n}} w_\mathrm{i}$. 
Until satisfactory computational quantifications are identified for additional metrics, our accuracy scoring is currently limited to the veracity score ($w_{\mathrm{ver}} = 1$).

\begin{figure}[bt]
    \centering
    \includegraphics[width=.5\textwidth]{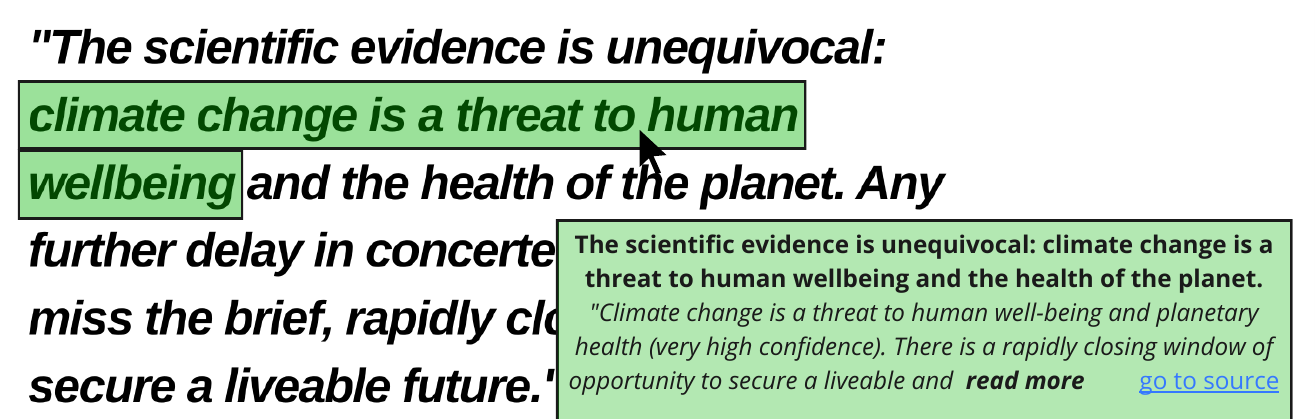}
    \caption[User interface mock-up showing the statement veracity score with color coding and ground truth reference.]{User interface mock-up showing the statement\footnotemark{} veracity score with color coding and ground truth reference.}
    \label{fig:UI}
\end{figure}

\secref{fig:UI} displays a mock-up of the user interface for a single statement.
Color-coded statements and source references are proposed to visually convey a numerical accuracy score.
A green highlighting indicating a confirmed statement, which can show additional details and refer to sources from the ground truth by hovering over it.
Even if the score is imperfect, referencing sources should make it more robust.

\addtocounter{footnote}{-1}

\section{Evaluation\label{sec:evaluation}}
To ensure that the methods and tools are up-to-date and properly contextualized, the approach and its implementation were evaluated through \textit{expert feedback} during intermediate presentations and concluding interviews with knowledge representation and processing experts from the ORKG team.
Additionally, a \textit{user survey} was conducted to assess the current state and inform future development.
\footnotetext{\url{https://science.nasa.gov/climate-change/effects/}}
The materials, such as the presentation and survey results, are available online\footnote{\url{https://github.com/borgnetzwerk/fact-checking/tree/main/Evaluation}}\footnote{\url{https://doi.org/10.5281/zenodo.17131832}}.

\subsection{Design\label{sec:design}}
\subsubsection{Expert Feedback}
The implementation and intermediate evaluation are aided by iterative communication with NLP experts, as described in \secref{sec:approach} and \ref{sec:implementation}.
Further presentations with the entire ORKG team, ranging from researchers to curators to software developers, collect and formalize this expert feedback.
One presentation is set in the middle of the development cycle, and one at the end.
The first presentation should detail the goals and current state of development of our workflow.
Alongside this explanation, eight questions are directed at the participants, such as 
\textit{"Q1 Is this approach up to date?"}, 
\textit{"Q4 The best way to extract triples is achieved through NER with a domain-specific model. Do you agree?"} and
\textit{"Q6 The path length is impactful to evaluate a claimed relation between A and B. Do you agree?"}.
These questions could be answered with \textit{"Yes / No / Other / No answer"} via polls.
The aim is to qualitatively discuss and consolidate the current findings and directions of the development process.
The results from these discussions are formalized into core statements and challenges for further refinement, to be presented to and confirmed by the team in a second iteration.

After familiarizing themselves with the goal and challenges of our approach, the participants are invited to individual interviews.
These online semi-structured interviews first show the presentation that the participants are already familiar with, asking for any further need of clarification or details.
Then, the questions and previously given answers are presented, asking for further explanation and a more detailed discussion of future suggestions.
These sessions aim to formalize deeper, more explicit expert perspectives.
Answering questions and participating in the discussions or interviews is always optional.

The analysis of the presentations is partially descriptive, with the given answers being counted, and partly coded, with the discussion points being transcribed and then mapped to common statements and challenges.
The analysis of the interviews was entirely coded, with recurring points adding up to a set of important statements and challenges.

\subsubsection{User Survey}
Over the weeks following the final presentation, an online survey is conducted. 
It is designed to present the findings of the work to an audience beyond the domain experts and captures a wider set of perspectives.
The survey ascertains whether users perceive scientific inaccuracy as a problem and a scientific accuracy score as a viable solution.
Next, it presents the current state of the tool using an example, and inquires in which circumstances and on which types of media the participants would utilize such a tool.
Participants could optionally disclose their age, expertise in suitable fields, and highest degree of education, allowing a clearer positioning of the survey sample relative to the general public.
The survey is distributed digitally with  broad inclusion criteria, as all consumers of online content are potential users.

\subsection{Results\label{sec:results}}
\subsubsection{Experts Feedback}
Both presentations had 27 participants and concluded in 15 to 20 minutes each, including discussions.
The refined insights provided by the experts where presented and 19 individuals voted at least once regarding their agreement to our findings.  
When presented with the identified core challenges and asked to select the two that, when overcome, would provide the greatest benefit, 15 individuals voted at least once here.
The statement and respective agreement, as well as the challenge votes, are presented in \secref{tab:presentation}.

\begin{table}[bth]
    \centering
    \caption{Expert feedback regarding agreement with our consolidated findings and their prioritization of future challenges}
    \label{tab:presentation}
    \begin{tabularx}{\linewidth}{X|c}
        \textbf{Concluding statement} & \textbf{\# Agree} \\
        \hline
        The optimal method for extracting triples is currently unclear. & 9 \\
        \hline
        LLMs have limitations and should not be used without proper checks to identify non-reproducible or hallucinated triples. & 16 \\
        \hline
        Scientific accuracy checks must take into account the context of statements, which cannot be represented by a single triple. & 13 \\
        \hline
        This tool is more likely to be used as an integrated rather than a stand-alone tool. & 4 \\
        \hline
        As long as "a perfect algorithm to check against the truth" is not achievable: Having an indication of what is more or less likely to be accurate is already helpful. & 8 \\
        \midrule
        \textbf{Challenge whose solution brings the greatest benefit} & \textbf{\# Votes} \\
        \midrule
        Handling the semantic alignment of natural language. & 6 \\
        \hline
        Handling LLM hallucinations and lacking reproducibility. & 9 \\
        \hline
        Fully automated triple extraction. & 0 \\
        \hline
        Keeping the context of statements, especially in empirical research. & 11 \\
        \hline
        Making statements about statements to show their confidence or time validity. & 1 \\
    \end{tabularx}
\end{table}

\begin{openboxwithtitle}{Most Important Challenges}
~Keeping statement context, especially in empirical research, followed by
handling LLM hallucinations and lacking reproducibility 
and the semantic alignment of natural language.
\end{openboxwithtitle}

Ten experts further provided detailed responses in sessions averaging 28 min (total: 4:42:00, max: 0:43:55, min: 0:13:39, $\sigma$: 0:10:40).
The gathered insight are detailed at length in Tremel~\cite[p.~50 ff.]{tremel_scientific_2024} and can be summarized as \secref{tab:statements}.
Nine of ten people interviewed agreed that they would use this tool,
just one was skeptical that the ground truth would be free of bias even when using published papers as source.
The participants generally found the approach to be up-to-date;
Yet, three participants deemed the question too vague, two of which stated that these technologies are older and are currently undergoing a renaissance.
It was noted by an expert with neurosymbolic AI experience that incoming claims data might be more relevant.
The same individual also suggested that graph walks could be examined in future research.
Two individuals referenced the employment of embeddings.
Five asserted that triple extraction constitutes a significant bottleneck in the approach.
Semantic parsing was mentioned explicitly by one individual.
A total of six individuals responded that LLMs demonstrate proficiency in particular tasks; however, they emphasized the necessity of exercising caution and implementing redundancy checks when employing these models.
Four individuals have noted the potential for a combination of NER and LLMs.
It was a consensus that LLMs possess numerous potential error sources.

\begin{table}[bth]
    \centering
    \caption{Mapped Expert Interview Statements}
    \begin{tabularx}{0.5\textwidth}{c|X}
        \textbf{ID} & \textbf{Statement} \\
        \midrule
        1 & Using the tool as a personal assistant can be helpful. It could communicate with a browser, a PDF reader, and other sources of information through a trusted graph.\\
        \hline
        2 & The use of LLMs and assembling information in knowledge graphs is often considered an up-to-date approach.\\
        \hline
        3 & When focusing on supporting claims, incoming data may be more relevant. Incoming nodes could be given more weight than outgoing nodes, similar to how SEO rankings work on websites.\\
        \hline
        4 & Graph walks can be utilized to match claims by expanding the connections of the trusted knowledge graph and counting the number of hops required to reach a similar triple to the one being checked.\\
        \hline
        5 & A common approach uses embeddings to compare the distances between different concepts.\\
        \hline
        6 & The bottleneck lies in the triple extraction phase. Semantic parsing presents a challenge due to the syntax, particularly when using RDF.\\
        \hline
        7 & There is a consensus that LLMs can be useful in specific domains. However, their effectiveness is debatable and depends on the requirements. Using them for independent tasks fast-tracks development.  Creating interfaces early on opens the possibility for more efficient and or reliable solutions for the subtask.\\
        \hline
        8 & A combination of NER and LLMs has potential. A mediator is necessary to ensure consistency in moderating statements or types of statements, particularly in cases involving trusted sources and popular media that require verification.\\
        \hline
        9 & LLMs have limitations and should not be used without proper checks to identify non-reproducible or hallucinated triples.
    \end{tabularx}
    \label{tab:statements}
\end{table}

\begin{openboxwithtitle}{Current State}
~\textbf{Knowledge graphs} are state-of-the-art interoperable knowledge representations.
When no exact match is found, embeddings or graph walks should be utilized, particularly with increased weighting of incoming edges.\\
\textbf{LLMs} are state-of-the-art for scalable statement extraction, but unreliable for semantic parsing and require the identification of non-reproducible or hallucinated triples.
NER mediators could complement LLMs to ensure consistency in moderating statements, instances, and classes.\\
\textbf{Our approach} is suitable as a personal assistant and should focus on the interoperability of tools.
\end{openboxwithtitle}

\subsubsection{User Survey}
\begin{figure}[bt]
    \centerline{\includegraphics[width=.5\textwidth]{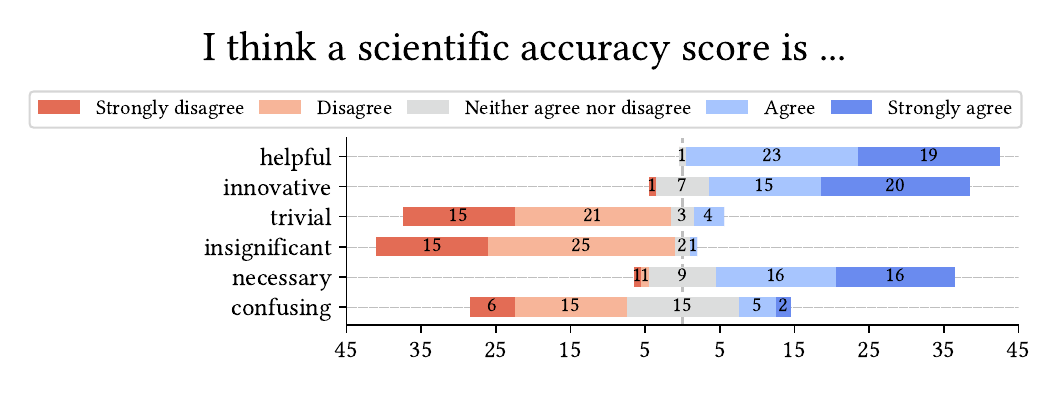}}
    \centerline{\includegraphics[width=.5\textwidth]{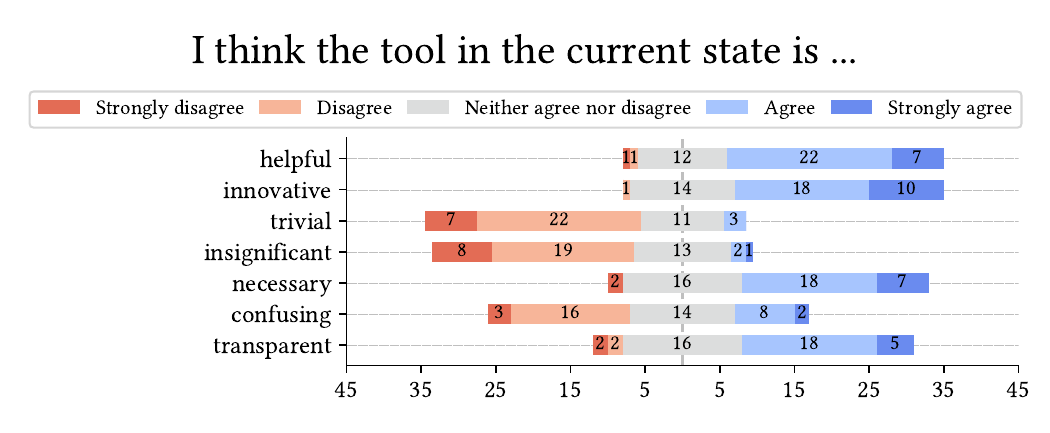}}
    \centerline{\includegraphics[width=.5\textwidth]{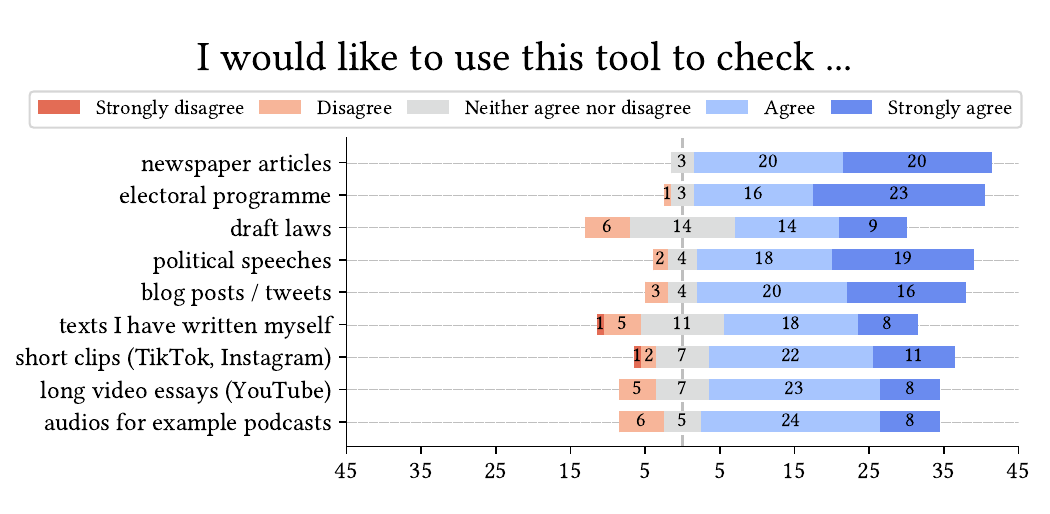}}
    \centerline{\includegraphics[width=.5\textwidth]{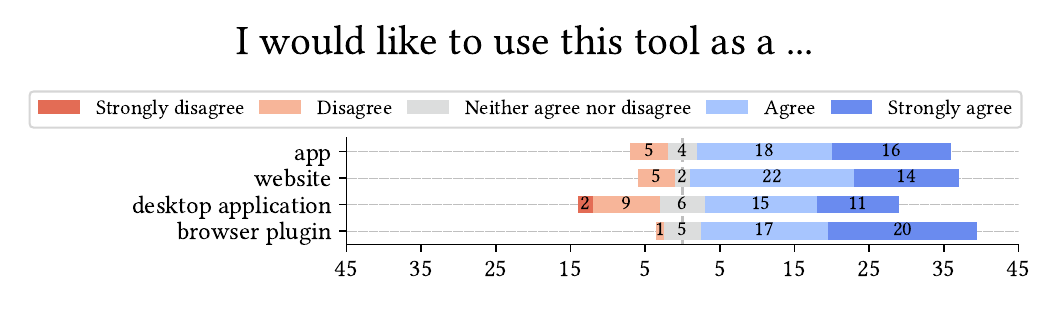}}
    \caption{User survey results (N=43).
    They show the demand for a scientific accuracy score, approve the tool in its current state, and indicate suitability for various media types.}
    \label{fig:user}
\end{figure}

The results from the 43 anonymous participants are depicted in \secref{fig:user}.
Participants primarily reported an age of 19-31 years and some educational degree (high school, bachelor or master's).
The results indicate the necessity and usability of the tool.
Notably, participants expressed interest in fact-checking a wide range of media formats, further supporting the demand for a versatile and scalable infrastructure in science communication.

\begin{openboxwithtitle}{User Perspective}
~A \textbf{Scientific Accuracy Score} and the \textbf{current implementation} are helpful and necessary.\\
All presented media require checking, most importantly electoral programs and newspaper articles, followed by political speeches and blog posts/tweets.\\
Browser plugin is the recommended form for the tool, followed by website and app.  
\end{openboxwithtitle}

\subsection{Threats to Validity\label{sec:validity}}
\subsubsection{Construct Validity}
The expert feedback presentation and interview were designed to identify computational fact-checking challenges.
Like the user survey, these measurements seek to inform and evaluate our development, focusing on relevant problems and the feasibility and usability of potential solutions to them.
This may unintentionally bias responses by suggesting desirable answers or lacking alternatives, which was sought to be mitigated by the "other" option and several ways to contribute feedback without a predefined structure, such as the discussions and interviews.
These subjective measures could be complemented with objective metrics, such as accuracy and recall, yet our work revealed that even subtasks, such as statement extraction, are not yet solved.
We focused on outlining challenges and solutions of the entire scope, while leaving modular optimization to specialized scientists already dedicated to advancing them.
To advance benchmarks, future work may test various modules with larger datasets and against comparative baselines and existing fact-checking tools.

\subsubsection{Internal Validity}
While the evaluation revealed positive user feedback and expert interest, it is difficult to determine whether these responses were influenced by the system quality or the presentation framing.
Some of these issues could be reduced by presenting to an audience with a scientific background in knowledge graphs and natural language processing, which are used to assess and evaluate the potential of new systems critically.
Yet, this introduces a bias, since the audience is familiar with the implications of contributing feedback to scientific evaluations.
This is mitigated not only by their respective scientific integrity but also by making any form of contribution optional, which is also reflected in the response rates.
All presentations, discussions, and interviews were semi-structured to balance consistency with room for custom contributions.
Beyond that, a more diverse audience with no prior background selection was similarly surveyed, revealing similar results.
The fact that these results are less technical is again compensated for by the prior expert presentations and interviews.
All participants were also reminded that critical feedback was welcome and important for improving the system.
Nevertheless, some social desirability bias or internal project alignment may have affected responses.

\subsubsection{External Validity}
During the development, groups and individuals were contacted, particularly most of the authors of the systems cited under related work, alongside many more finally not included in the further development.
Conclusively, the chance that major systems, approaches, or ground truth knowledge graphs were missed is reduced, yet it remains.
The user sample is also limited to individuals within reach of the authors.
While the survey is publicly shared and re-sharing was particularly encouraged, these methods generally reach few beyond second or third degree contacts.
This is visible due to the academic overrepresentation and poses limitations in generalizing findings to the general public or non-academic media consumers.
Expanding the survey distribution to more demographically varied and international audiences in future iterations would help address this limitation.

\subsubsection{Conclusion Validity}
Although the expert interviews and survey results were largely consistent, the sample size and qualitative nature may reduce the robustness of the conclusions.
Due to these limitations, no statistical significance tests were applied to survey results, and coding of expert statements may have introduced interpretative bias.
To enhance reliability, interview and survey data are made public in the repository and thesis, as far as possible.
The highlighted key insights were also always raised by multiple individuals in separate sessions, reducing the relative influence bias.
Still, caution is warranted when extrapolating the identified challenges and recommendations beyond the scope of this study.

\section{Discussion and Future Work\label{sec:discussion}}
We demonstrated a workflow to semantically align a ground truth alongside media statements to facilitate computational fact-checking.
While promising in addressing misinformation, this approach revealed several issues not currently solved by state-of-the-art computational fact-checking:
Fact-checking is always only as good as its facts, but the availability of high-quality FAIR~\cite{wilkinson_fair_2016} ground truth knowledge graphs is limited.
Especially for urgent and cohesive domains such as climate change, larger-scale semantification collaboration is required before a comprehensive corpus can be built.
When facing these scalability challenges and limited number of ground truth triples, it is worthwhile to pay close attention to approaches proposed by communities such as \#semanticClimate.

Knowledge graphs have been demonstrated to be effective tools for consistently and reproducibly verifying statements based on prior knowledge. 
The open world assumption can complicate the process of falsifying statements, where the utilization of graph-based scoring has been identified as a potential solution.
The validity of path length as a measure of veracity is skewed by the graph granularity and the existence of contradictory semantic information, which may be partitioned into separate, homogeneously modeled graphs.
Providing evidence and references for claims can, to a certain extent, hold more significance than the score itself.
Knowledge graphs possess the capacity to efficiently evolve over time, thereby enhancing the efficacy of subsequent fact-checking processes.
Potentially, federating multiple graphs and sources with different publication dates would allow a broader, up-to-date observation of scientific knowledge.
Natural language processing enables the extraction of information on a large scale.
Given the considerable reliance on LLMs and their limitations, requiring ongoing monitoring of model accuracy, dedicated future work is required.
As outlined in the \nameref{sec:validity}, the ongoing advances in the field of LLM and semantification, such as fine-tuning for triple extraction, are immediately beneficial to the effectiveness of our approach.
In addition, it is critical to closely monitor the substantial energy consumption associated with training LLMs and utilizing them.
Rebound effects~\cite{thiesen_rebound_2008} can nullify any efficiency gains that result from increased usage.
While the knowledge graph ground truth representation scales well, the current statement extraction approaches do not, requiring significant future work.

Beyond these issues, even a veracity score devoid of errors cannot fully address the issue of misinformation.
Three further challenges must be addressed:
context, acceptance, and integration into the public discourse.
\textit{Firstly,} hyperfocusing factuality as a parameter could create other blind spots, such as context (e.g. sarcastic), clearness (e.g. convoluted), authenticity (e.g. cited incorrectly), subjectivity (e.g. ``I feel like''), or confidence (e.g. ``likely'').
Even if a statement cannot be proven factually wrong (such as any subjective statement starting with ``I feel like ...''), it can strongly influence a discussion.
Similarly, a factually correct statement taken out of context can misrepresent a complex topic.
Consequently, the present scoring implementation must be expanded to include the previously discussed score categories.
\textit{Secondly,} it must be noted that citizens who harbor distrust toward the scientific community will not consent to a ground truth that has been constructed upon scientific principles. 
Consequently, these citizens are likely to reject this fact-checking method.
The social and behavioral sciences are better equipped to address this challenge.
At the same time, this approach can assist communities within the public discourse who are interested in incorporating scientific knowledge and collectively advancing society.
\textit{Thirdly,} extensive manual fact-checking resources exist, yet reaching their intended audience remains an equally challenging issue in the attention economy.
This issue should be addressed by prioritizing the fact-checking of media that is widely viewed and establishing fact-checking scores in a prominent and accessible manner.

Evidently, computational fact-checking to alleviate misinformation requires significant further future work, including on:
A knowledge community that crowdsources ground truth synthesis using FAIR digital infrastructure, extensive popular media annotation, and effective result dissemination.

\begin{openboxwithtitle}{Research Finding}
~Natural language processing and knowledge graphs can help quantify the scientific accuracy of secondary literature by semantifying its content and matching it against ground truth representations.
Yet, scaling is inhibited by a lack of FAIR climate change ground truth and unreliable semantification.\\
\textbf{Future work} needs to collaboratively advance three interconnected inhibitors:
\textbf{FAIR ground truth curation}, particularly in critical knowledge infrastructure;
\textbf{semantification accuracy}, particularly tripple extraction;
\textbf{energy-efficient reuse}, particularly regarding AI-accelerated rebound effects.
\end{openboxwithtitle}

\section{Conclusion\label{sec:conclusion}}
This work presented a novel workflow to assess the scientific accuracy of online media using a neurosymbolic approach to computational fact-checking. 
By combining LLM-based statement extraction with knowledge-graph-based reasoning, the system offers a promising approach to the growing challenge of misinformation, particularly in the context of climate change. 
The prototype implementation demonstrates the feasibility of automating parts of the fact-checking process and provides a scientific accuracy score that can support users in critically evaluating media claims.

However, the effectiveness of such computational approaches remains fundamentally constrained by the availability and quality of machine-readable, FAIR-compliant ground truth knowledge. 
Our evaluation with domain experts and a broader user survey indicates both strong interest and practical relevance. 
However, it also reveals key technical and infrastructural limitations, especially in statement context preservation, semantic alignment, and the handling of LLM-generated artifacts.

To fully realize this approach's potential, future work must focus on scaling and enriching climate knowledge graphs, improving semantic interoperability, and developing mechanisms for transparent, explainable scoring. 
Equally important is the need to address public trust and media literacy, as fact-checking tools must not only be technically robust but also socially acceptable and ethically aligned. 
By fostering interdisciplinary collaboration and an open knowledge infrastructure, we can advance towards a more resilient and scientifically grounded public discourse.

\bibliographystyle{ieeetr}
\bibliography{main}

\end{document}